\documentclass[letterpaper,journal]{IEEEtran}
\usepackage{amsmath,amssymb}
\usepackage{array}
\usepackage{booktabs}
\usepackage{makecell}
\usepackage{stfloats}
\usepackage{url}
\usepackage{graphicx}
\usepackage{cite}
\usepackage{placeins}
\usepackage[hidelinks]{hyperref}

\newcommand{\authorbiography}[3]{
  \IfFileExists{#2}{
    \begin{IEEEbiography}[{\includegraphics[width=1in,height=1.25in,clip,keepaspectratio]{#2}}]{#1}
      #3
    \end{IEEEbiography}
  }{
    \begin{IEEEbiographynophoto}{#1}
      #3
    \end{IEEEbiographynophoto}
  }
}

\begin{document}

\title{ReLoco: Unified Humanoid Locomotion and Fall Recovery via Behavior-Conditioned Motion Priors}

\author{Lequn Fu$^{1}$, Yufeng Hu$^{1}$, Yijun Zhong$^{1}$, Jie Zhang$^{1}$, Xiao Li$^{1}$, Lingxiang Ji$^{1}$, Zhiyuan Xu$^{2}$, \mbox{Jian Tang$^{2}$,~\emph{Fellow, IEEE}}, and Shiqi Li$^{1,\dagger}$\thanks{Lequn Fu, Yufeng Hu, Yijun Zhong, Jie Zhang, Xiao Li, Lingxiang Ji and Shiqi Li are with the $^{1}$Huazhong University of Science and Technology, Wuhan 430074, China (e-mail: fulq@hust.edu.cn; huyufeng@hust.edu.cn; zhongyijun@hust.edu.cn; d202480318@hust.edu.cn; li\_xiao@hust.edu.cn; m202570806@hust.edu.cn; sqli@hust.edu.cn).}\thanks{Zhiyuan Xu and Jian Tang are with the $^{2}$Beijing Innovation Center of Humanoid Robots, Beijing 100086, China (e-mail: eric.xu@x-humanoid.com; jian.tang@x-humanoid.com).}\thanks{$^{\dagger}$Corresponding author.}}

\maketitle

\begin{abstract}
Humanoid robots must recover from falls and resume commanded motion to sustain locomotion. Learning these behaviors in one policy requires supervision that accommodates both upright movement and ground-contact recovery. We present ReLoco, a framework that uses behavior-conditioned adversarial motion priors to learn locomotion, get-up from arbitrary fallen postures, and command resumption within one proprioceptive policy. During training, robot posture selects the expert pool and conditions a shared discriminator with two behavior-specific outputs. Task rewards and shared regularization support command tracking and recovery, while temporary assistance facilitates exploration. A simulation-based refinement stage supplies robot-executed motion references for this supervision. Ablations examine conditional supervision, output separation, and reference quality under matched settings. At a common training iteration, the full policy achieves $96.80\%$ independent get-up success and $96.96\%$ push-recovery success, including command resumption. Its torso planar velocity root-mean-square error is $0.1266$~m/s, $14.7\%$ lower than the conditional single-output variant, with similar joint-angle discrepancy measured by dynamic time warping. Deployment on the 64.5-kg Tiangong~3.0 humanoid demonstrates walking, running, get-up, and locomotion resumption after falls using one proprioceptive actor, without controller switching or on-robot policy fine-tuning.
\mbox{Project page: \url{https://lequn-f.github.io/ReLoco}}
\end{abstract}


\section{Introduction}
\label{sec:introduction}

A humanoid robot that falls during locomotion must regain support, stand up, and continue its task. These abilities require coordinated motion across markedly different postures and contacts. Walking emphasizes command following and upright support, whereas recovery can involve rolling, arm support, and substantial torso rotation. A velocity command specifies the motion to be resumed, but does not describe how to establish support from a fallen posture. A single policy must therefore learn recovery actions that restore support before returning to velocity tracking.

Legged control has advanced through model-based planning~\cite{Meduri2023biconmp}, combined planning and learned tracking~\cite{Jenelten2024dtc}, and reinforcement learning (RL) in parallel simulation~\cite{Makoviychuk2021isaacgym,Rudin2022walk}. Learned controllers have achieved adaptive bipedal locomotion~\cite{Kumar2022adapting}, real-world humanoid walking~\cite{Radosavovic2024humanoid}, versatile dynamic gaits~\cite{Li2024cassie}, and compliant sim-to-real control~\cite{Masuda2024simtoreal}. Recent approaches further address difficult terrain, policy smoothness, and sparse footholds~\cite{Cui2025terrain,Chen2025smooth,Wang2025beamdojo}. Sustained mobility additionally requires recovery when disturbances take the robot outside its nominal locomotion states.

Human motion provides useful structure for whole-body coordination. Reference-tracking methods learn physically simulated skills from prescribed trajectories~\cite{Peng2018deepmimic,Fussell2021supertrack,Luo2023phc}. Adversarial motion priors (AMP) learn a motion reward from examples~\cite{Peng2021amp}, while skill embeddings and conditional models support diverse behaviors~\cite{Peng2022ase,Tessler2023calm,Won2022physicsvae}. Human-reference and generative-prior approaches extend motion supervision to humanoid locomotion~\cite{Tang2024humanmimic,Zhang2024whole,Zhang2025gmp}. Related systems demonstrate expressive and agile whole-body control~\cite{Cheng2024exbody,He2025hover,He2025asap,Li2025amo}, human-to-humanoid teleoperation~\cite{He2024h2o,Fu2025humanplus,He2025omnih2o}, and contextual skills learned from video~\cite{Allshire2025videomimic}. Retargeting addresses embodiment differences in human-to-robot transfer~\cite{Colmenares2024retargeting} and semantic consistency across characters~\cite{Zhang2024smt}; recent frameworks further support humanoid retargeting and motion imitation~\cite{Araujo2026gmr,Chen2025motionimitation}.

Recovery has also become a learned robot capability. HoST addresses standing-up exploration, optimization, and deployable motion through a force curriculum, multiple critics, and motion constraints~\cite{Huang2025host}. Other learning-based recovery methods include HumanUP, HiFAR, and FRASA~\cite{He2025humanup,Chen2025hifar,Gaspard2025frasa}. AHC consolidates behavior-specific policies through distillation and RL fine-tuning~\cite{Zhao2026ahc}, while X-Loco trains a vision-based generalist through specialist supervision~\cite{Wang2026xloco}. These studies motivate learning recovery together with the motion that the robot must resume.

After a disturbance, the robot's configuration depends on the preceding gait, disturbance direction, and sequence of contacts, and need not coincide with a reference frame. Recovery must also begin from the velocity and support conditions produced by the fall, which can differ from those in a demonstration. AMP guides locally observed motion without prescribing frame-by-frame reference alignment~\cite{Peng2021amp}, allowing the feedback policy to respond from the state actually reached.

We present ReLoco (Recovery and Locomotion), a framework for unified locomotion and fall recovery through behavior-conditioned motion supervision. A posture-derived supervision route selects the locomotion or get-up reference pool. The discriminator scores motion histories together with their recorded conditions, using a shared feature network and two behavior-specific outputs. Posture-appropriate task rewards support command following or getting up, while shared penalties regulate actuation and joint motion across both routes. This formulation accommodates recovery-specific contacts and postures while retaining the velocity command that the robot must eventually track.

To provide suitable references for this supervision, we refine retargeted human motions in simulation. Kinematic retargeting preserves human coordination but can introduce contact and joint-limit errors on the robot. An offline tracking policy refines locomotion and get-up motions under the contact and actuator models used for final policy training. The resulting robot-executed examples supply the expert data used to train the final controller through adversarial motion supervision.

Recovery training covers configurations that need not align with individual reference frames. Reference-state initialization exposes the policy to fallen and transitional postures, while velocity perturbations generate falls during motion. Temporary assistance supports early recovery learning and is removed as training progresses. These settings provide experience for learning the transition back to commanded locomotion. Deployment uses a proprioceptive actor with velocity commands, without the reference tracker, recovery assistance, or an externally supplied behavior label.

We evaluate the resulting policy through torso planar velocity tracking, motion similarity, independent get-up, and command resumption after disturbances. AMP ablations compare pooled and conditioned supervision; a conditional single-output comparison holds expert sampling and task objectives fixed to assess output separation. Reference-refinement experiments examine the effect of expert-motion quality. Independent get-up trials use settled fallen states, whereas disturbances during locomotion test recovery from moving states. Hardware tests on the 64.5-kg Tiangong~3.0 humanoid demonstrate walking, running, get-up from arbitrary fallen postures, and locomotion resumption after falls. The main contributions are:
\begin{enumerate}
  \item A behavior-conditioned motion-prior scheme that selects reference pools using robot posture and scores motion histories with their conditions, supporting command tracking, fall recovery, and resumption of commanded motion within the same actor.
  \item A simulation-based refinement procedure that converts retargeted locomotion and get-up motions into robot-executed references, reducing contact and joint-limit violations in the expert data used for adversarial supervision.
  \item Demonstration of locomotion, autonomous get-up from arbitrary fallen postures, and resumption of the commanded motion after falls on a full-size humanoid using the same deployed proprioceptive policy.
\end{enumerate}

\section{Method}
\label{sec:method}

\begin{figure*}[t]
  \centering
  \includegraphics[width=\textwidth]{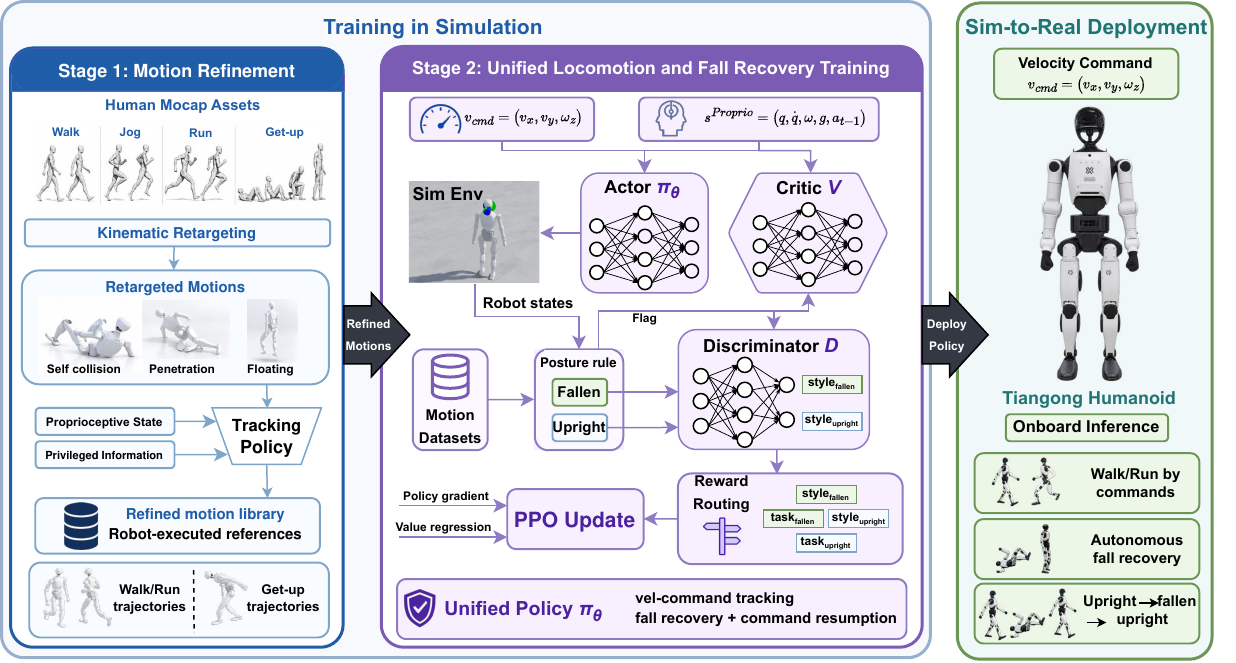}
  \caption{Overview of ReLoco. Retargeted human motions are refined in simulation to form locomotion and get-up reference sets. Posture conditions expert sampling and the output of a shared discriminator. Together with task rewards and common regularization, this supervision trains one actor for locomotion and recovery.}
  \label{fig:overview}
\end{figure*}

ReLoco trains one proprioceptive actor for locomotion, fall recovery, and resumption of commanded motion through behavior-conditioned AMP supervision. Robot posture selects the expert pool and conditions a shared discriminator with two outputs. Task rewards, shared regularization, and a recovery curriculum support policy optimization. Offline refinement of human motions supplies the robot-executed locomotion and get-up references used for this supervision. The discriminator and route labels are used only during training. Fig.~\ref{fig:overview} summarizes the framework.

\subsection{Problem Formulation}
\label{sec:formulation}

We model the simulated robot and environment as a Markov decision process $(\mathcal{S},\mathcal{A},\mathcal{P},r,\gamma)$ and train an asymmetric actor--critic. The actor receives proprioception, velocity commands, and the previous action; the critic additionally receives training-only state information (Table~\ref{tab:policy_observations}(a)). At each control step, the actor samples $\mathbf{a}_t\sim\pi_\theta(\cdot\mid\mathbf{o}_t)$. The action specifies an offset from the default joint configuration:
\begin{equation}
  \mathbf{q}^{d}_t=\mathbf{q}^{\mathrm{default}}+\mathbf{s}\odot\mathbf{a}_t,
  \label{eq:action_mapping}
\end{equation}
The joint targets are tracked by a low-level PD controller:
\begin{equation}
  \boldsymbol{\tau}^{\mathrm{cmd}}_t=K_p(\widetilde{\mathbf{q}}^{d}_t-\mathbf{q}_t)-K_d\dot{\mathbf{q}}_t,
  \label{eq:pd_control}
\end{equation}
where $\mathbf{s}$ contains the action scales and $\widetilde{\mathbf{q}}^{d}_t$ is the delayed target used by the actuator model.

During training, the binary supervision-route label $b_t\in\{0,1\}$ selects the training objectives. For policy samples, $b_t=0$ denotes upright standing or locomotion, and $b_t=1$ denotes non-upright motion, including falling and recovery. The label is available to the critic but excluded from the actor observation; it is not a deployment control mode.

\begin{table}[!t]
    \centering
    \caption{Policy observations and training configuration.}
    \label{tab:policy_observations}
    \label{tab:network_config}
    \label{tab:domain_rand}
    \footnotesize
    \setlength{\tabcolsep}{3pt}
    \renewcommand{\arraystretch}{1.06}
    \begin{tabular*}{\columnwidth}{@{\extracolsep{\fill}}lcc@{}}
        \toprule
        \multicolumn{3}{@{}l}{\textit{(a) Policy observations}} \\
        \textbf{Observation term} & \textbf{Actor} & \textbf{Critic} \\
        \midrule
        Linear velocity              & --           & $\checkmark$ \\
        Angular velocity             & $\checkmark$ & $\checkmark$ \\
        Projected gravity            & $\checkmark$ & $\checkmark$ \\
        Velocity command             & $\checkmark$ & $\checkmark$ \\
        Joint positions              & $\checkmark$ & $\checkmark$ \\
        Joint velocities             & $\checkmark$ & $\checkmark$ \\
        Previous action              & $\checkmark$ & $\checkmark$ \\
        Supervision-route label $b_t$ & --          & $\checkmark$ \\
        \bottomrule
    \end{tabular*}
    \par\smallskip
    \begin{tabular*}{\columnwidth}{@{\extracolsep{\fill}}llc@{}}
        \toprule
        \multicolumn{3}{@{}l}{\textit{(b) Network architectures}} \\
        \textbf{Network} & \textbf{Hidden layers / output} & \textbf{Activation} \\
        \midrule
        Actor & $512\!\rightarrow\!256\!\rightarrow\!128\!\rightarrow\!23$ & ELU \\
        Critic & $512\!\rightarrow\!256\!\rightarrow\!128\!\rightarrow\!1$ & ELU \\
        Discriminator & $1024\!\rightarrow\!512\!\rightarrow\!256$; two heads & ReLU \\
        \bottomrule
    \end{tabular*}
    \par\smallskip
    \begin{tabular*}{\columnwidth}{@{\extracolsep{\fill}}lcc@{}}
        \toprule
        \multicolumn{3}{@{}l}{\textit{(c) Domain randomization}} \\
        \textbf{Parameter} & \textbf{Sampling range} & \textbf{Perturbation} \\
        \midrule
        Restitution coefficient & $[0.0,0.5]$ & Absolute \\
        Base mass & $[-3.0,3.0]$ kg & Additive \\
        Limb mass & $[0.8,1.2]\times$ & Scaling \\
        Base/torso CoM & $[-0.03,0.03]$ m & Additive \\
        Joint armature & $[0.9,1.1]\times$ & Scaling \\
        \bottomrule
    \end{tabular*}
\end{table}

During training, noise is applied to the actor proprioception, while critic inputs remain noise-free.

\subsection{Behavior-Conditioned Supervision}
\label{sec:behavior_conditioned_supervision}

Posture identifies the reference motions and immediate task objectives used for supervision. All samples update the same actor, with joint-motion and actuation penalties active in both routes.

\subsubsection{Posture-Based Supervision Routing}
\label{sec:condition_router}

We infer the supervision label from torso height $h_t$ and vertical alignment $u_t=\mathbf{e}_z^\top R_t^{\mathrm{torso}}\mathbf{e}_z$:
\begin{equation}
  b_t=\begin{cases}0,&h_t\geq h_{\mathrm{loc}}\ \land\ u_t\geq u_{\mathrm{loc}},\\1,&\text{otherwise},\end{cases}
  \label{eq:binary_stage}
\end{equation}
where $h_{\mathrm{loc}}=0.9$~m and $u_{\mathrm{loc}}=0.7$. Standing, walking, and running share the upright route. These thresholds route training supervision and differ from the recovery-success criteria. The condition is the one-dimensional binary input $\mathbf c_t=[b_t]$, which is not empirically normalized. Non-upright samples use the recovery route.

\subsubsection{Behavior-Conditioned Adversarial Motion Prior}
\label{sec:cond_amp}

The motion prior should guide upright coordination during locomotion and ground-contact motion during recovery. Each frame contains 76 motion features $\boldsymbol\psi_t=[\boldsymbol\rho_t,{}^a\mathbf v_t,{}^a\boldsymbol\omega_t,\mathbf q_t,\dot{\mathbf q}_t,{}^a\mathbf p_t^{\mathrm{key}}]$: yaw-removed torso orientation in a six-dimensional tangent--normal representation, torso-frame linear and angular velocities (three dimensions each), 23 joint positions and 23 joint velocities, and 18 torso-relative position coordinates for two trunk bodies, the two ankles, and the two elbows. The torso's own relative position is zero and is retained. Appending $\mathbf c_t$ gives 77 inputs per frame. A policy window contains ten consecutive frames, $\mathbf Y_t=(\boldsymbol\psi_{t-9},\ldots,\boldsymbol\psi_t)$, with their saved conditions $\mathbf C_t=(\mathbf c_{t-9},\ldots,\mathbf c_t)$. The shared feature network receives all ten motion--condition pairs flattened into a $10\times77=770$-dimensional vector. Windows crossing a posture threshold retain their historical conditions and enter the policy replay buffer when AMP is active. The final-frame label $b_t$ additionally selects the output head.

We maintain a locomotion library $\mathcal E_{\mathrm u}$ and a get-up library $\mathcal E_{\mathrm{rec}}$. The binary label saved in the rollout selects the expert pool:
\begin{equation}
  \mathbf Y^E\sim
  \begin{cases}
    \mathcal E_{\mathrm u}, & b_t=0,\\
    \mathcal E_{\mathrm{rec}}, & b_t=1.
  \end{cases}
  \label{eq:conditioned_expert_sampling}
\end{equation}
Each expert window takes ten consecutive frames from one motion file at a randomly sampled start time. Its condition is the file's pool label at every frame, rather than a label recomputed from expert posture. Thus, upright terminal frames in a get-up file retain the recovery label.

A feature network $F_{\phi_F}$ is shared by two scalar output heads $H_{\phi_0}$ and $H_{\phi_1}$:
\begin{equation}
  D_{\phi,b_t}(\mathbf Y_t,\mathbf C_t)
  =H_{\phi_{b_t}}\!\left(F_{\phi_F}(\mathbf Y_t,\mathbf C_t)\right),
  \label{eq:binary_disc_heads}
\end{equation}
where $\phi=\{\phi_F,\phi_0,\phi_1\}$. The separate outputs assign a score for each behavior while retaining a common motion representation. Let $b^\pi$ denote a policy window's final-frame label and $b^E$ an expert window's source-pool label. The selected scores are $d^\pi=D_{\phi,b^\pi}(\mathbf Y^\pi,\mathbf C^\pi)$ and $d^E=D_{\phi,b^E}(\mathbf Y^E,\mathbf C^E)$. Following least-squares AMP~\cite{Peng2021amp}, the discriminator learns targets of $-1$ for policy windows and $+1$ for expert windows:
\begin{align}
  \mathcal L_D={}&\tfrac12\mathbb E_\pi[(d^\pi+1)^2]
  +\tfrac12\mathbb E_E[(d^E-1)^2] \nonumber\\
  &+\lambda_{\mathrm{gp}}\mathcal R_{\mathrm{gp}}.
  \label{eq:disc_loss}
\end{align}
Here $\mathcal R_{\mathrm{gp}}$ denotes gradient regularization on expert samples. Updates through either head train the shared feature network. The selected policy score gives the style reward
\begin{equation}
  r_t^{\mathrm{style}}=\beta\Delta t\max\!\left(0,1-\tfrac14(D_{\phi,b_t}(\mathbf Y_t^\pi,\mathbf C_t)-1)^2\right).
  \label{eq:style_reward}
\end{equation}

\subsubsection{Task Rewards and Policy Optimization}
\label{sec:reward_routing}

Task rewards complement the motion prior: upright rewards encourage commanded locomotion, while recovery rewards encourage the robot to raise its torso. Let $\mathcal R_{\mathrm{u}}$, $\mathcal R_{\mathrm{n}}$, and $\mathcal R_{\mathrm{s}}$ contain the upright, non-upright, and shared terms, respectively. The task reward is
\begin{equation}
  \begin{aligned}
    r_t^{\mathrm{task}}=\Delta t\Bigg[&(1-b_t)\sum_{i\in\mathcal R_{\mathrm{u}}}w_i r_{i,t}
    +b_t\sum_{i\in\mathcal R_{\mathrm{n}}}w_i r_{i,t}\\
    &+\sum_{i\in\mathcal R_{\mathrm{s}}}w_i r_{i,t}\Bigg].
  \end{aligned}
  \label{eq:task_reward_routing}
\end{equation}
Waist-joint deviation, torque, acceleration, action-rate, and joint-limit penalties belong to the shared reward and remain active in both supervision routes. Table~\ref{tab:reward_terms} lists all task-reward terms and weights; $\Delta t=0.02$~s.

For the table, $T_{\mathrm{ds}}=\min_{i\in\mathcal F}T_i^c$ is the shorter foot-contact duration, and $F_b^{\max}=\max_{\tau}\|\mathbf F_{b,\tau}\|_2$ is the peak contact force on body $b$. We use $\Delta\psi_i=\operatorname{wrap}(\psi_i-\psi_a)$ for foot orientation, $\eta_{\mathrm{feet}}=\sum_{i\in\mathcal F}(|\phi_i|+\chi_{\mathrm{nt}}|\Delta\psi_i|)$, and $\ell_{\mathrm{lim}}(\mathbf q)=\sum_j([q_j^{\min}-q_j]^+ + [q_j-q_j^{\max}]^+)$ for soft joint limits. The standing indicator is $s_t=\mathbf 1[\|\mathbf v^{\mathrm{cmd}}_{xy}\|_2<0.1\land|\omega^{\mathrm{cmd}}_z|<0.1]$. For velocity tracking, $\rho(\mathbf e;s_t)=\exp(-4\|\mathbf e\|_1)$ under standing commands and $\exp(-4\|\mathbf e\|_2^2)$ otherwise.

\begin{table}[t]
    \centering
    \caption{Task-reward terms and shared penalties.}
    \label{tab:reward_terms}
    \footnotesize
    \setlength{\tabcolsep}{5.0pt}
    \renewcommand{\arraystretch}{1.04}
    \begin{tabular}{@{}lll@{}}
        \toprule
        \textbf{Reward Term} & \textbf{Equation} & \textbf{Weight} \\
        \midrule
        \multicolumn{3}{l}{\textit{Upright Task Rewards}} \\[0.25ex]
        Planar Velocity
        & $\rho(\mathbf v^{\mathrm{cmd}}_{xy}-{}^{Y}\mathbf v^{a}_{xy};s_t)$
        & $1.2$ \\
        Yaw Velocity
        & $\rho(\omega^{\mathrm{cmd}}_z-\omega^a_z;s_t)$
        & $1.2$ \\
        Hard Velocity Tracking
        & $r_{\mathrm{track,hard}}$
        & $0.5$ \\
        Torso Orientation
        & $\exp[-10(|\phi_a|+|\theta_a|)]$
        & $0.8$ \\
        \makecell[l]{Prolonged Double\\Support}
        & $\chi_{\mathrm{ls}}\mathbf 1[T_{\mathrm{ds}}>0.5\,\mathrm{s}]$
        & $-100.0$ \\
        Feet Orientation
        & $\exp(-2\eta_{\mathrm{feet}})$
        & $0.2$ \\
        Foot Stumble
        & $\mathbf 1[\exists i:\|\mathbf F_i^{xy}\|_2>5|F_i^z|]$
        & $-2.0$ \\
        Feet Sliding
        & $\sum_{i\in\mathcal F}c_i\|\mathbf v^W_{i,xy}\|_2$
        & $-0.25$ \\
        Undesired Contacts
        & $\sum_{b\notin\mathcal F}\mathbf 1[F_b^{\max}>1\,\mathrm{N}]$
        & $-1.0$ \\
        Stand-Still Pose
        & $s_t\|\mathbf q-\mathbf q^0\|_1$
        & $-0.3$ \\
        \midrule
        \multicolumn{3}{l}{\textit{Non-Upright Task Rewards}} \\[0.25ex]
        Torso Height
        & $r_{\mathrm{height}}(h_a;h^*=1.261\,\mathrm{m})$
        & $1.0$ \\
        Upward Velocity Limit
        & $[v_z^a-1.1]^+$
        & $-5.0$ \\
        \midrule
        \multicolumn{3}{l}{\textit{Shared Rewards and Regularization}} \\[0.25ex]
        Waist-Joint Deviation
        & $\|\mathbf q_{\mathcal J_w}-\mathbf q^0_{\mathcal J_w}\|_1$
        & $-0.1$ \\
        Feet Separation
        & $[0.2-d_y]^+ + [d_y-0.8]^+$
        & $-2.0$ \\
        Termination
        & $\mathbf 1_{\mathrm{terminated}}$
        & $-200.0$ \\
        All-Joint Deviation
        & $\|\mathbf q-\mathbf q^0\|_1$
        & $-0.05$ \\
        Joint Torque
        & $\|\boldsymbol\tau\|_2^2$
        & $-1.0\!\times\!10^{-6}$ \\
        Joint Acceleration
        & $\|\ddot{\mathbf q}\|_2^2$
        & $-2.5\!\times\!10^{-7}$ \\
        Action Rate
        & $\|\mathbf a_t-\mathbf a_{t-1}\|_2^2$
        & $-0.05$ \\
        Joint Soft Limits
        & $\ell_{\mathrm{lim}}(\mathbf q)$
        & $-2.0$ \\
        \bottomrule
    \end{tabular}
\end{table}

In Table~\ref{tab:reward_terms}, $a$ denotes the torso, $Y$ its yaw-aligned frame, $\mathcal F$ the two feet, $c_i$ the foot-contact indicator, and $d_y$ the lateral separation of the feet. During recovery, the torso-height reward is
\begin{equation}
  r_{\mathrm{height}}=\exp\!\left[-\frac{(h_a-1.261)^2}{0.5^2}\right].
  \label{eq:height_reward}
\end{equation}
To sharpen velocity tracking during locomotion, we use planar speed error $e_v=\|\mathbf v^{\mathrm{cmd}}_{xy}-{}^{Y}\mathbf v^{a}_{xy}\|_2$ and yaw-rate error $e_\omega=|\omega^{\mathrm{cmd}}_z-\omega^a_z|$. The unweighted reward is
\begin{equation}
  r_{\mathrm{track,hard}}=\chi_{\mathrm{cmd}}\!\left[\tfrac12(e^{-10e_v}+e^{-10e_\omega})-0.2(e_v+e_\omega)\right],
  \label{eq:hard_tracking_reward}
\end{equation}
Here $\chi_{\mathrm{cmd}}=\mathbb I[\|\mathbf v^{\mathrm{cmd}}_{xy}\|_2>0\ \lor\ |\omega^{\mathrm{cmd}}_z|>0]$ disables this term for zero commands. The height and hard-tracking weights are $1.0$ and $0.5$. The indicators $\chi_{\mathrm{ls}}$ and $\chi_{\mathrm{nt}}$ refer to low-speed and non-turning conditions. Under standing commands, $\chi_{\mathrm{ls}}=0$, so prolonged double support is not penalized. Here $[x]^+=\max(x,0)$.

AMP supervision is active for sufficiently large motion commands and for non-upright states. With activation indicator $g_t$, PPO~\cite{Schulman2017ppo} uses
\begin{equation}
 r_t=\begin{cases}
  0.5r_t^{\mathrm{task}}+0.5r_t^{\mathrm{style}}, & g_t=1,\\
  r_t^{\mathrm{task}}, & g_t=0.
 \end{cases}
 \label{eq:combined_reward}
\end{equation}
Shared regularization remains in the task component. Returns and advantages are calculated from the combined reward. For policy optimization, route-wise advantage normalization standardizes upright and non-upright samples separately:
\begin{equation}
  \widehat A_t=\frac{A_t-\mu_{b_t}}{\sigma_{b_t}+10^{-8}},
  \label{eq:route_advantage_normalization}
\end{equation}
where $\mu_b$ and $\sigma_b$ are the mean and standard deviation of advantages for samples assigned to supervision route $b$. Both routes participate in the same policy training process. Raw rewards and critic return targets remain unchanged.

\subsection{Simulation-Refined Motion References}
\label{sec:motion_prior_construction}

Kinematic retargeting can preserve a human motion's overall shape while leaving errors in robot contacts and joint coordination. We use self-collected human motions and selected sequences from BONES-SEED~\cite{BonesStudioSeed}. We retarget them using GMR~\cite{Araujo2026gmr} and execute the trajectories with a tracking policy in simulation. Selected robot rollouts form the expert references:
\begin{equation}
  \mathcal{M}_{\mathrm{human}}\xrightarrow{\mathcal{R}_{\mathrm{kin}}}\mathcal{M}_{\mathrm{raw}}\xrightarrow{\pi_{\mathrm{track}},\,\mathcal{P}}\mathcal{M}_{\mathrm{ref}}.
  \label{eq:motion_refinement_pipeline}
\end{equation}
The tracking policy maximizes
\begin{equation}
  J_{\mathrm{refine}}=\mathbb{E}\!\left[\sum_{t=0}^{T-1}\gamma^t(r_t^{\mathrm{track}}+r_t^{\mathrm{reg}})\right],
  \label{eq:motion_refinement_objective}
\end{equation}
The tracking reward $r_t^{\mathrm{track}}$ keeps the simulated motion close to its target, while $r_t^{\mathrm{reg}}$ discourages undesirable contacts and abrupt joint motion. Table~\ref{tab:refinement_rewards} gives the reward terms.

\begin{table}[t]
    \centering
    \caption{Reward terms for reference-motion refinement.}
    \label{tab:refinement_rewards}
    \footnotesize
    \setlength{\tabcolsep}{2.2pt}
    \renewcommand{\arraystretch}{1.05}
    \begin{tabular}{@{}lll@{}}
        \toprule
        \textbf{Reward Term} & \textbf{Equation} & \textbf{Weight} \\
        \midrule
        \multicolumn{3}{l}{\textit{Tracking Rewards} ($r_{\mathrm{track}}$)} \\[0.3ex]
        Global Torso Pos.
        & $\exp(-\|\mathbf p_{\mathrm{torso}}-\hat{\mathbf p}_{\mathrm{torso}}\|^2/\sigma_p^2)$
        & $0.5$ \\
        Global Torso Ori.
        & $\exp(-\|\mathbf R_{\mathrm{torso}}\ominus\hat{\mathbf R}_{\mathrm{torso}}\|^2/\sigma_R^2)$
        & $0.5$ \\
        Relative Body Pos.
        & $\exp(-\|\mathbf p_{\mathrm{body}}-\hat{\mathbf p}_{\mathrm{body}}\|^2/\sigma_{\mathrm{b,pos}}^2)$
        & $1.0$ \\
        Relative Body Ori.
        & $\exp(-\|\mathbf R_{\mathrm{body}}\ominus\hat{\mathbf R}_{\mathrm{body}}\|^2/\sigma_{\mathrm{b,ori}}^2)$
        & $1.0$ \\
        Body Linear Vel.
        & $\exp(-\|\mathbf v_{\mathrm{body}}-\hat{\mathbf v}_{\mathrm{body}}\|^2/\sigma_v^2)$
        & $1.0$ \\
        Body Angular Vel.
        & $\exp(-\|\boldsymbol\omega_{\mathrm{body}}-\hat{\boldsymbol\omega}_{\mathrm{body}}\|^2/\sigma_\omega^2)$
        & $1.0$ \\
        \midrule
        \multicolumn{3}{l}{\textit{Regularization Penalties} ($r_{\mathrm{reg}}$)} \\[0.3ex]
        Joint Soft Limits
        & $\sum_j([q_j^{\min}-q_j]^+ + [q_j-q_j^{\max}]^+)$
        & $-2.0$ \\
        Undesired Contacts
        & $\sum_{b\notin\mathcal B_{\mathrm{allowed}}}\mathbf 1_{\mathrm{contact}}(b)$
        & $-0.1$ \\
        Action Rate ($L_2$)
        & $\|\mathbf a_t-\mathbf a_{t-1}\|_2^2$
        & $-0.05$ \\
        Joint Torque
        & $\|\boldsymbol\tau\|_2^2$
        & $-1.0\!\times\!10^{-6}$ \\
        Joint Accel. ($L_2$)
        & $\sum_j\ddot q_j^{\,2}$
        & $-2.5\times10^{-7}$ \\
        \bottomrule
    \end{tabular}
\end{table}

Refinement uses the robot and actuator model employed for policy training. The squared-torque, joint-acceleration, action-rate, and soft-limit penalties have the same weights as in the final policy. The refinement objective additionally penalizes contacts outside the allowed foot and hand set.

The refined trajectories form the locomotion and get-up libraries, supplying robot-executed references without adding a tracking controller to deployment.

\subsection{Training Strategy}
\label{sec:training_strategy}

Recovery training covers varied initial postures and transitions following a loss of balance. Reference-state initialization broadens posture coverage, velocity perturbations induce falls, and temporary assistance supports early exploration.

\textit{Reference-state initialization.} At reset, half of the environments start from randomly chosen get-up frames, ranging from fallen to nearly upright. The other half sample from the full library of locomotion and recovery motions. We initialize the robot's global pose and velocity, together with joint positions and velocities, from the selected frames.

\textit{Push disturbance randomization.} Training also includes translational and yaw-velocity perturbations. They occur independently in each environment, with intervals sampled from $[15,20]$~s. The offsets are drawn from
\begin{equation}
  \begin{aligned}
    \Delta v_x&\sim\mathcal{U}(-5,5)~\mathrm{m/s}, & \Delta v_y&\sim\mathcal{U}(-3,3)~\mathrm{m/s},\\
    \Delta\omega_z&\sim\mathcal{U}(-4,4)~\mathrm{rad/s}. &&
  \end{aligned}
  \label{eq:push_randomization}
\end{equation}
These planar and yaw-velocity offsets can cause loss of balance and complete falls.

\textit{Recovery assistance curriculum.} External assistance has been used to facilitate humanoid standing-up exploration~\cite{Huang2025host}. In our training setup, an upward force is applied to the robot's head while it is non-upright:
\begin{equation}
  \mathbf f_{t,\mathrm{head}}^{\mathrm{assist}}=F_k\mathbf e_z\,\mathbb I[b_t=1],\qquad F_k\in[0,F_0],
  \label{eq:recovery_assistance}
\end{equation}
Here $F_0$ is the initial force and $F_k$ its current value.
After the windowed get-up success rate exceeds $0.8$, assistance decreases by $20$~N at each reset from a fallen state until it reaches zero. Evaluation and hardware deployment use no assistance.

\subsection{Sim-to-Real Transfer}
\label{sec:simtoreal}
Robot and actuator dynamics are identified following PACE~\cite{Bjelonic2026pace}. Domain randomization covers uncertainty in contact, inertia, and actuation parameters (Table~\ref{tab:domain_rand}(c)), with actuator-command delay randomized separately.

Training uses these calibrated and randomized dynamics. Deployment runs the same proprioceptive actor throughout locomotion and recovery, without the reference tracker, discriminator, assistance force, or supervision router.
\section{Experiments}

\label{sec:experiments}

We evaluate how the unified policy combines commanded locomotion and fall recovery. Reference-refinement experiments assess the effect of expert-motion quality, while the AMP ablation distinguishes the joint effect of conditioning and expert routing from the additional effect of output separation. Recovery-only comparisons and hardware trials assess autonomous get-up and continuation of the commanded task after disturbances.

\subsection{Experimental Setup}

\label{sec:setup}

The Tiangong~3.0 humanoid is 1.70~m tall and weighs 64.5~kg. It has 31 articulated degrees of freedom (DoF), of which 23 are controlled by the policy; the two head joints and three wrist DoF on each arm are excluded. We train policies in Isaac Lab with proximal policy optimization (PPO)~\cite{Schulman2017ppo}. Comparisons within each test use the same robot model, action interface, and evaluation states. Table~\ref{tab:network_config}(b) gives the network structures.

Undisturbed locomotion is evaluated using torso planar velocity root-mean-square error (RMSE), joint-angle discrepancy measured by dynamic time warping (DTW), and a contact-based foot-sliding proxy. Random single-axis commands use $v_x^{\mathrm{cmd}}\in[-0.8,-0.5]\cup[0.5,2.5]$~m/s or $v_y^{\mathrm{cmd}}\in[-0.8,-0.5]\cup[0.5,0.8]$~m/s, with zero yaw command. For this test, RMSE pools both command axes over a fixed 5-s measurement window following a 2-s warm-up. RMSE and sliding use the common trials that remain upright throughout ($h_t\geq0.9$~m and $u_t\geq0.7$). The sliding proxy time-averages the sum of contact-masked horizontal ankle-link center-of-mass speeds. DTW measures the mean absolute joint-angle discrepancy along the optimal alignment path for 3-s trajectories of the 23 controlled joints, averaged equally over 28 selected straight-motion training references, including mirrors.

Independent get-up and recovery-only comparisons use the same posture-based success criterion: torso height $h_t>1.0$~m and uprightness $u_t>0.8$ must be reached within 5~s and maintained for a further 0.5~s. Push recovery applies the same criterion following a confirmed grounded fall and additionally requires command resumption. For this post-fall task-resumption test, a 5-s transition after upright recovery is followed immediately by a fixed 5-s evaluation window. The torso planar velocity RMSE over this window must satisfy
\begin{equation}
  e_v^{\mathrm{RMSE}}=
  \sqrt{\frac{1}{N}\sum_{k=1}^{N}
  \left\|\mathbf v^{\mathrm{torso}}_{xy,k}
  -\mathbf v^{\mathrm{cmd}}_{xy}\right\|_2^2}
  \leq0.3~\mathrm{m/s},
  \label{eq:task_resumption}
\end{equation}
where $N$ is the number of samples in the fixed evaluation window.

Recovery evaluations use no automatic resets, early termination, or recovery assistance, and all trials enter the success-rate denominators. The recovery-only comparison additionally reports recovery time, absolute joint mechanical work $E=\sum_t\sum_j|\tau_{t,j}\dot q_{t,j}|\Delta t$, peak torque, and mean absolute joint acceleration. Hardware outcomes distinguish continued locomotion, recovery followed by task resumption, and protection events.

\subsection{Simulation-Based Reference-Motion Refinement}

\label{sec:exp_motion_refinement}

Kinematic retargeting provides reference poses but does not ensure that a robot can execute them. We compare 43 matched sequences before and after simulation-based refinement. The raw set $\mathcal{M}_{\mathrm{raw}}$ contains the retargeted motions; $\mathcal{M}_{\mathrm{ref}}$ contains their refined, robot-executed counterparts. Table~\ref{tab:motion_refinement_data} reports mean $\pm$ standard deviation across the 43 sequences.

\begin{figure}[t]
  \centering
  \includegraphics[width=\columnwidth]{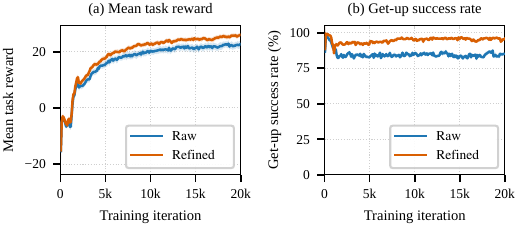}
  \caption{Effect of reference-motion refinement on AMP training. (a) Mean task reward. (b) Get-up success rate. Raw uses the original retargeted motions; Refined uses simulation-refined references. Shading denotes within-run variation over $20{,}000$ training iterations.}
  \label{fig:motion_refinement_training}
\end{figure}

Refinement lowers all three mean constraint-violation rates to below $1\%$. Mean body-position jerk also decreases by $35.5\%$. To test whether cleaner references help learning, we train the same AMP setup once with raw references and once with refined references. Fig.~\ref{fig:motion_refinement_training} compares the first $20{,}000$ iterations.

At $20{,}000$ iterations, refined references yield a mean task reward of approximately $25.5$, compared with $22.0$ for raw motions, and logged get-up success of about $95\%$ versus $85\%$. Together with the lower reference-violation rates, these results support using robot-executed motions as AMP references. The comparison uses one training run per condition.

\begin{table}[t]
  \centering
  \caption{Reference-motion quality (43 matched motions, mean$\pm$standard deviation).}
  \label{tab:motion_refinement_data}
  \footnotesize
  \setlength{\tabcolsep}{5pt}
  \begin{tabular}{@{}lcc@{}}
    \toprule
    \textbf{Metric} & \textbf{Raw} & \textbf{Refined} \\
    \midrule
    Ground penetration (\%) $\downarrow$ & $28.99\pm27.60$ & $0.75\pm2.09$ \\
    Self-collision (\%) $\downarrow$ & $12.91\pm14.77$ & $0.35\pm1.27$ \\
    Joint-limit violation (\%) $\downarrow$ & $12.90\pm17.24$ & $0.06\pm0.38$ \\
    Body-position jerk (m/s$^3$) $\downarrow$ & $406.88\pm310.44$ & $262.49\pm118.73$ \\
    \bottomrule
  \end{tabular}
\end{table}

\subsection{Unified Locomotion and Fall Recovery in Simulation}

\label{sec:exp_simulation}

\textbf{Ablation of behavior-conditioned AMP supervision.} Table~\ref{tab:supervision_ablation} compares Task-only RL, Unconditional AMP, Single-head conditional AMP, and Ours. Task-only RL omits AMP rewards and discriminator updates. Unconditional AMP pools the references without route conditioning. Both conditional variants use the same route-conditioned histories and expert sampling, differing in one versus two discriminator outputs. Task-reward routing, regularization, actor observations, and route-wise advantage normalization remain fixed. AMP variants equally weight task and style rewards when AMP is active; Task-only RL uses the full task reward. The unconditional-to-conditional comparison evaluates conditioning and routed sampling jointly.

Independent get-up tests use matched randomized settled fallen states and zero commands. Push-recovery tests apply a 1600--2000~N horizontal torso force for 0.3~s in a random direction during forward walking, running, backward, or lateral motion. All variants are evaluated at iteration $20{,}000$, using one original training run per variant.

\begin{figure}[t]
  \centering
  \includegraphics[width=\columnwidth]{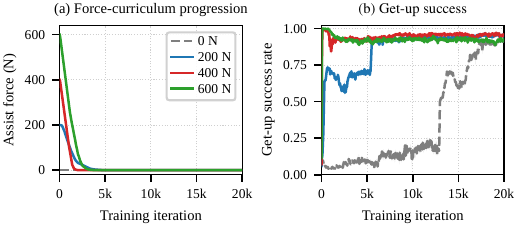}
  \caption{Initial recovery assistance. (a) Assistance-force progression.
  (b) Recorded get-up success. Final rates follow assistance withdrawal;
  results are from individual training runs.}
  \label{fig:force_curriculum}
\end{figure}

\begin{table*}[t]
  \centering
  \caption{Ablation results at training iteration $20{,}000$. Metrics are mean $\pm$ sample standard deviation over five paired evaluation batches, with 250 trials per batch in each recovery test. Push-recovery SR requires upright recovery and command tracking under the RMSE criterion in Eq.~\eqref{eq:task_resumption}. $I_{90}$ is the logged training milestone.}
  \label{tab:supervision_ablation}
  \footnotesize
  \setlength{\tabcolsep}{2.4pt}
  \renewcommand{\arraystretch}{1.07}
  \begin{tabular}{@{}lcccccc@{}}
    \toprule
    \textbf{Method}
    & \shortstack{\textbf{Torso planar RMSE}\\(m/s)$\downarrow$}
    & \shortstack{\textbf{Joint DTW}\\(rad)$\downarrow$}
    & \shortstack{\textbf{Foot-sliding proxy}\\(m/s)$\downarrow$}
    & \shortstack{\textbf{Get-up SR}\\(\%)$\uparrow$}
    & \shortstack{\textbf{Push-recovery SR}\\(\%)$\uparrow$}
    & \shortstack{$I_{90}$\\(iterations)$\downarrow$} \\
    \midrule
    Task-only RL
    & $0.0717\pm0.0006$ & $0.2944\pm0.0005$ & $0.0635\pm0.0025$
    & $0.00\pm0.00$ & $8.24\pm0.92$ & $>20{,}000$ \\
    Unconditional AMP
    & $0.8599\pm0.0554$ & $0.1651\pm0.0002$ & $0.0956\pm0.0025$
    & $94.64\pm0.92$ & $92.32\pm1.25$ & $4{,}063$ \\
    Single-head conditional AMP
    & $0.1485\pm0.0015$ & $0.1303\pm0.0003$ & $0.0878\pm0.0019$
    & $90.16\pm2.48$ & $93.36\pm1.19$ & $6{,}357$ \\
    \textbf{Ours}
    & $0.1266\pm0.0009$ & $0.1307\pm0.0002$ & $0.0856\pm0.0019$
    & $96.80\pm1.47$ & $96.96\pm0.73$ & $3{,}600$ \\
    \bottomrule
  \end{tabular}
\end{table*}

\begin{figure*}[t]
  \centering
  \includegraphics[width=\textwidth]{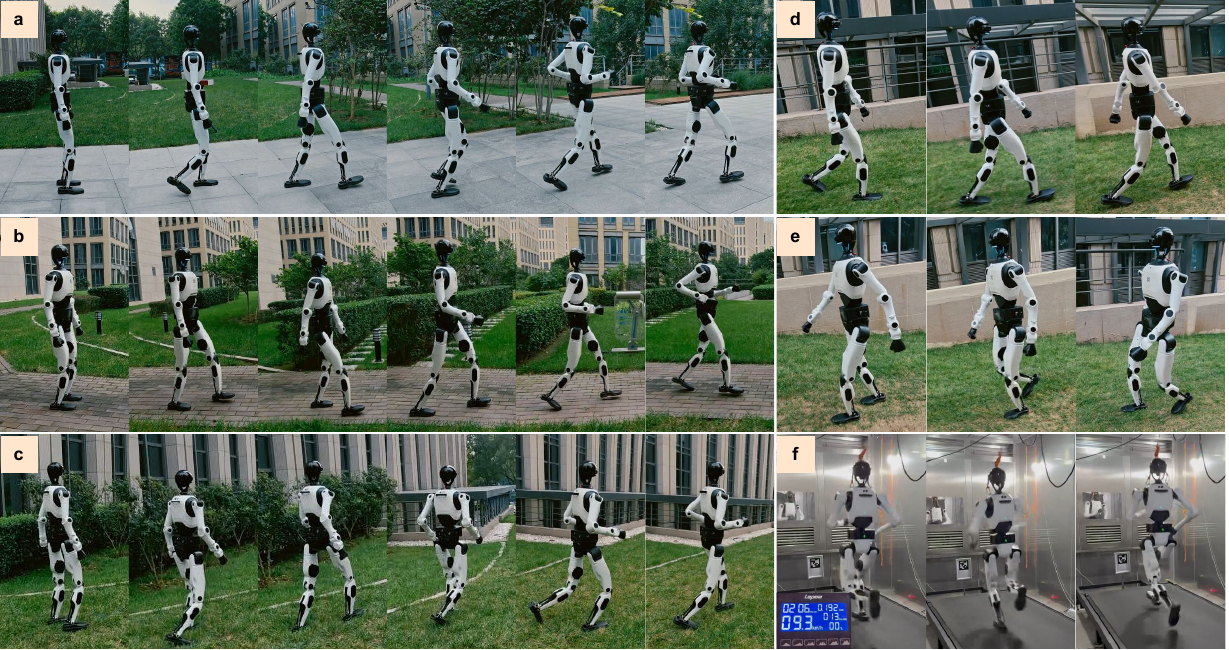}
  \caption{Real-world locomotion on (a) stone-tiled pavement, (b) brick pavement, (c) grass, (d) a downhill grassy slope, (e) an uphill grassy slope, and (f) a treadmill. Standing, walking, and running are generated by the same policy without explicit gait-mode switching; the treadmill belt-speed setting reaches $9.3$~km/h.}
  \label{fig:real_locomotion}
\end{figure*}

Task-only RL achieves the lowest tracking error and foot-sliding proxy but fails all independent get-up trials. Unconditional AMP achieves high recovery success with substantially higher tracking error. Both conditional variants have lower tracking RMSE and joint-angle DTW than Unconditional AMP. The single-output variant, however, has lower independent get-up success than Unconditional AMP (Table~\ref{tab:supervision_ablation}).

With the conditional setup held fixed, separate outputs further reduce torso planar velocity RMSE by $14.7\%$ and increase get-up success by $6.64$ percentage points. Push-recovery success, including command resumption, improves by $3.60$ percentage points, with comparable joint-angle DTW.

The training logs show that the policy learns sustained unassisted recovery earlier. We define $I_{90}$ as the ending iteration of the first 500 consecutive iterations with logged recovery success $\geq90\%$ and zero logged mean assistance force. This milestone uses the original training metric, separately from the evaluation criteria. Ours reaches it at iteration 3600, compared with 6357 for Single-head conditional AMP, requiring $43.4\%$ fewer iterations in these runs. Task-only RL does not reach this milestone by iteration $20{,}000$.

\textbf{Recovery capability of the unified policy.} We compare our unified controller with HoST~\cite{Huang2025host} and a BeyondMimic-style tracker trained on one prescribed get-up sequence~\cite{liao2026beyondmimic} (Table~\ref{tab:recovery_baselines}). Each controller is tested on the same robot from 1000 reference-pose starts and 1000 matched randomized settled fallen states, with a 400-Nm actuator torque limit. Randomized states are obtained by releasing the unactuated robot from random orientations and allowing it to settle under gravity. The independent get-up ablation uses the same success criterion but a different set of 1250 randomized states, which can yield a different success rate.

The unified policy retains high get-up success from randomized fallen states: $98.3\%$, compared with $99.9\%$ for HoST and $49.7\%$ for the single-reference tracker (Table~\ref{tab:recovery_baselines}). HoST recovers faster ($1.57$ versus $2.02$~s), whereas our policy has lower mechanical work ($3.1$ versus $10.7$~kJ), peak torque, and mean absolute joint acceleration on each method's successful trials. These results characterize the recovery-speed and actuation-cost balance of a controller that also performs locomotion.

\begin{table}[t]
  \centering
  \caption{Recovery-only comparison under matched initial conditions and
  success criteria. Success rates use all 1000 episodes per condition;
  other entries are mean $\pm$ standard deviation over successful episodes.}
  \label{tab:recovery_baselines}
  \scriptsize
  \setlength{\tabcolsep}{2pt}
  \renewcommand{\arraystretch}{1.08}
  \resizebox{\columnwidth}{!}{\begin{tabular}{@{}lccccc@{}}
    \toprule
    \textbf{Method} &
    \shortstack{\textbf{Success}\\(\%) $\uparrow$} &
    \shortstack{\textbf{Time}\\(s) $\downarrow$} &
    \shortstack{\textbf{Mech. work}\\(kJ) $\downarrow$} &
    \shortstack{\textbf{Peak $|\tau|$}\\(Nm) $\downarrow$} &
    \shortstack{\textbf{Mean $|\ddot{\mathbf q}|$}\\(rad/s$^2$) $\downarrow$} \\
    \midrule
    \multicolumn{6}{l}{\emph{Reference-motion start pose}} \\
    HoST~\cite{Huang2025host}
      & 99.3 & $1.64\pm0.45$ & $11.2\pm2.4$ & $399\pm4$ & $139\pm21$ \\
    Single-ref. tracker~\cite{liao2026beyondmimic}
      & 100 & $3.84\pm0.00$ & $1.7\pm0.0$ & $400\pm0$ & $17\pm1$ \\
    \textbf{Ours}
      & 100 & $2.32\pm0.45$ & $2.3\pm0.6$ & $270\pm69$ & $36\pm5$ \\
    \midrule
    \multicolumn{6}{l}{\emph{Randomized settled fallen states}} \\
    HoST
      & 99.9 & $1.57\pm0.75$ & $10.7\pm5.2$ & $400\pm2$ & $131\pm21$ \\
    Single-ref. tracker
      & 49.7 & $2.93\pm1.15$ & $1.8\pm1.9$ & $343\pm77$ & $20\pm14$ \\
    \textbf{Ours}
      & 98.3 & $2.02\pm0.65$ & $3.1\pm0.8$ & $307\pm52$ & $42\pm7$ \\
    \bottomrule
  \end{tabular}}
\end{table}

\textbf{Force curriculum analysis.} We test initial upward assistance forces of $F_0\in\{0,200,400,600\}$~N using the withdrawal rule in Section~\ref{sec:training_strategy}. Fig.~\ref{fig:force_curriculum} shows individual runs over 20,000 iterations with 4096 parallel environments (approximately $1.97\times10^9$ environment steps). Final success statistics use a 50-iteration window after assistance reaches zero.

Without assistance, recorded success first exceeds $0.8$ near $16.9$k iterations. With $400$~N, the force reaches zero near $1.9$k iterations, and the final unassisted rate is $0.955\pm0.005$, compared with $0.915\pm0.004$ without assistance. Increasing the initial force to $600$~N delays withdrawal to about $3.7$k iterations and yields a final rate of $0.920\pm0.009$. The $400$-N setting gives the highest final unassisted rate among these runs and is used in the remaining experiments.

\begin{figure*}[t]
  \centering
  \includegraphics[width=\textwidth]{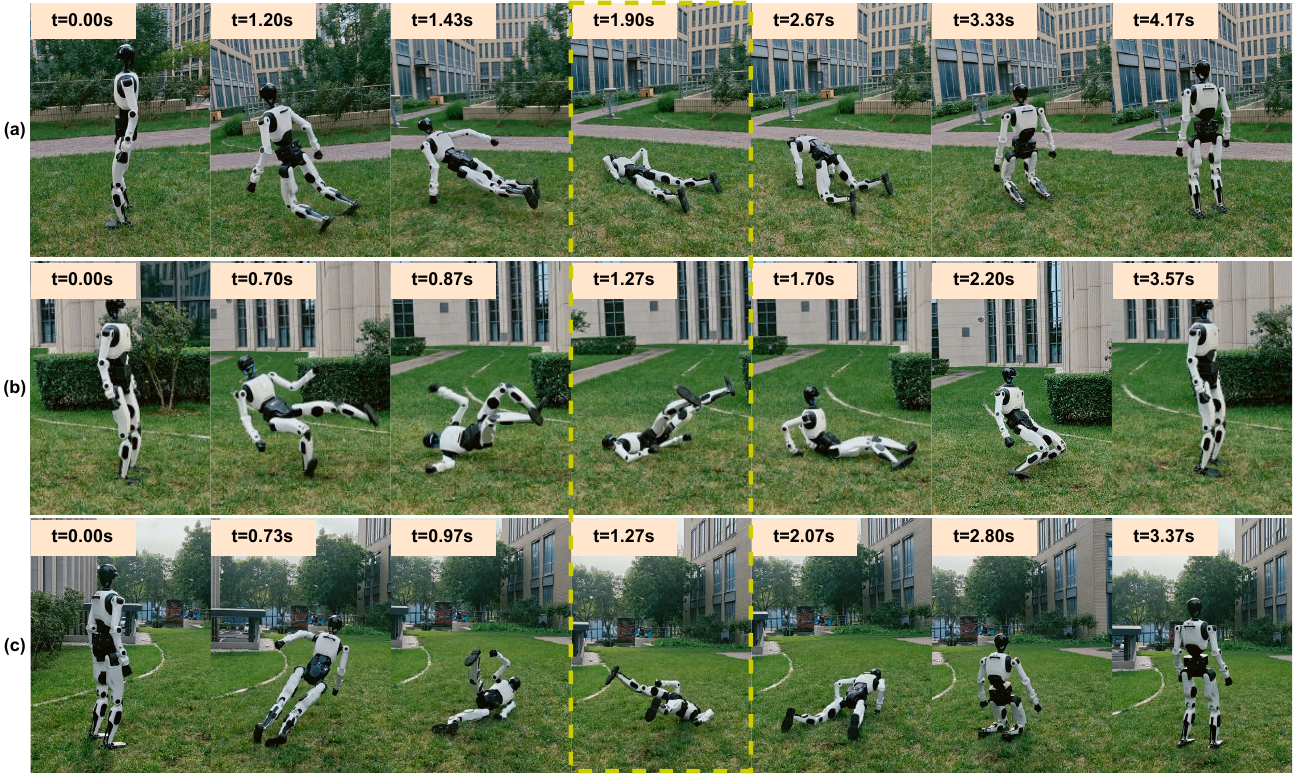}
  \caption{Autonomous recovery following external kicks from (a) rear, (b) front, and (c) side directions. Yellow dashed boxes highlight representative postural adjustments during the fall-to-recovery transition.}
  \label{fig:real_fall_recovery}
\end{figure*}

\subsection{Real-Robot Experiments}

\label{sec:exp_real_robot}

We transfer the trained policy directly to Tiangong~3.0, without on-robot policy fine-tuning or a separate recovery controller. We evaluate locomotion on varied surfaces, autonomous get-up from arbitrary fallen postures, and continuation or resumption of walking and running after disturbances.

\subsubsection{Locomotion Across Real-World Terrains}

Fig.~\ref{fig:real_locomotion} shows the robot on tiled and brick pavement, grass, uphill and downhill grassy slopes, and a treadmill. Across these settings, the same policy produces standing, walking, and running without an externally selected gait mode.

For walking, commanded forward, lateral, and yaw velocities cover $v_x^c\in[-0.8,1.6]$~m/s, $v_y^c\in[-0.8,0.8]$~m/s, and $\omega_z^c\in[-3,3]$~rad/s. Commands are sampled separately along each axis; when forward motion and turning are combined, $|\omega_z^c|\leq1$~rad/s. For running, only forward velocity is commanded, with targets from $1.6$ to $2.5$~m/s and zero lateral and yaw commands. Running emerges around a forward command of $1.6$~m/s.

The robot completed approximately 10~min of randomized-command outdoor locomotion in total without a fall. A treadmill demonstration used belt-speed settings up to $9.3$~km/h.

\subsubsection{Recovery Across Fallen Configurations}

We applied external kicks while the robot was standing (Fig.~\ref{fig:real_fall_recovery}). Of 47 attempts, 34 resulted in falls after which the policy remained active, five did not produce a fall, and eight triggered immediate hardware protection. The policy recovered autonomously in all 34 trials where control continued, covering prone, supine, and side-lying configurations. Table~\ref{tab:real_locomotion_disturbance}(a) includes every attempt.

Timing statistics use the first ten recoveries in each posture category. Four further recoveries are included in the overall outcomes. Fall duration runs from the kick to the established fallen configuration, and get-up duration from that configuration to stable upright recovery.

\begin{table}[!tb]
  \centering
  \caption{Hardware disturbance outcomes and recovery timing.
  (a) All recorded attempts, including protection events; locomotion recoveries
  resumed the commanded task. No additional task failures were observed.
  (b) Timing over the first ten successful recoveries per posture
  (30 trials total), reported as mean $\pm$ standard deviation.}
  \label{tab:real_fall_recovery}
  \label{tab:real_locomotion_disturbance}
  \footnotesize
  \setlength{\tabcolsep}{3pt}
  \renewcommand{\arraystretch}{1.07}
  \begin{tabular*}{\columnwidth}{@{\extracolsep{\fill}}lcccc@{}}
    \toprule
    \multicolumn{5}{@{}l}{\textit{(a) All recorded disturbance attempts}} \\
    \textbf{Condition} & \shortstack{\textbf{All}\\\textbf{attempts}}
    & \shortstack{\textbf{No}\\\textbf{fall}}
    & \shortstack{\textbf{Fall and}\\\textbf{recovery}}
    & \shortstack{\textbf{Hardware}\\\textbf{protection}} \\
    \midrule
    Standing & 47 & 5 & 34 & 8 \\
    Walking & 10 & 9 & 1 & 0 \\
    Running & 11 & 7 & 3 & 1 \\
    \bottomrule
  \end{tabular*}
  \par\smallskip
  \begin{tabular*}{\columnwidth}{@{\extracolsep{\fill}}lccc@{}}
    \toprule
    \multicolumn{4}{@{}l}{\textit{(b) Balanced recovery-timing subset}} \\
    \textbf{Posture} & \textbf{Success} & \textbf{Fall (s)} & \textbf{Get-up (s)} \\
    \midrule
    Prone      & 10/10 & $0.78\pm0.06$ & $3.04\pm0.73$ \\
    Supine     & 10/10 & $1.38\pm0.42$ & $2.07\pm0.11$ \\
    Side-lying & 10/10 & $0.71\pm0.06$ & $2.81\pm0.46$ \\
    \midrule
    Overall (30 trials) & 30/30 & $0.96\pm0.39$ & $2.64\pm0.64$ \\
    \bottomrule
  \end{tabular*}
\end{table}

Mean get-up time in the balanced 30-trial subset is $2.64\pm0.64$~s (Table~\ref{tab:real_fall_recovery}(b)). Supine recovery is fastest at $2.07\pm0.11$~s, followed by side-lying at $2.81\pm0.46$~s and prone at $3.04\pm0.73$~s. The adjustments highlighted in Fig.~\ref{fig:real_fall_recovery} show that the policy can reposition the body before standing up.

\begin{figure}[!t]
  \centering
  \includegraphics[width=\columnwidth]{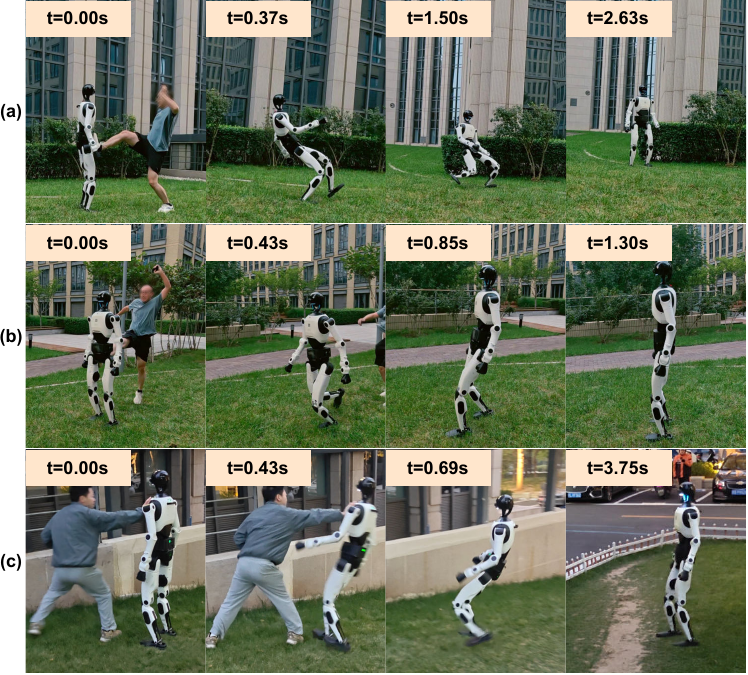}
  \caption{Real-world balance recovery after manual disturbances without a complete ground-level fall. (a) and (b) Responses on approximately level grassy terrain. (c) Response to a push on an inclined grassy surface, showing postural adjustment and restoration of an upright stance. The slope sequence is a qualitative demonstration outside the 20 level-ground locomotion trials.}
  \label{fig:real_disturbance}
\end{figure}

\subsubsection{Disturbance Response During Locomotion}

We next disturbed the robot during commanded walking and forward running. Ten evaluable trials were recorded for each gait on level ground. Manual kicks varied in direction and magnitude, but the applied forces were not measured. We classified each response as continued locomotion without falling, get-up followed by task resumption, or failure to resume. An immediate impact-protection event was recorded separately when the policy could not be evaluated further.

The robot maintained or resumed its commanded motion in all 20 evaluable trials (Table~\ref{tab:real_locomotion_disturbance}(a)). Sixteen continued without falling: nine walking and seven running. Four completed the full locomotion--fall--recovery--locomotion sequence: one walking and three running. One additional running kick triggered immediate hardware protection. These trials provide hardware examples of task continuation and post-fall locomotion resumption.

Fig.~\ref{fig:real_disturbance} shows balance recovery without a full fall on approximately level grass (a,b) and after a push on a grassy slope (c). The slope sequence is separate from the 20 level-ground trials. Since kick and push magnitudes were not controlled, these demonstrations do not establish a maximum recoverable force or a quantitative terrain-robustness comparison.

\subsection{Discussion}

Task recovery requires the robot to stand up and resume the commanded motion. The push-recovery tests therefore require renewed command tracking after standing up, and hardware trials demonstrate locomotion resumption after falls.

Tracking accuracy and recovery capability must be considered together when evaluating unified control. Task-only RL tracks velocity well but fails the independent get-up tests, while pooled AMP combines high recovery success with substantially higher tracking error. Conditioning and routed sampling improve tracking and reference similarity; output separation adds recovery gains within that setup (Table~\ref{tab:supervision_ablation}). Reference refinement further supports this learning process by reducing violations in the expert motions.
\FloatBarrier
\section{Conclusion}
\label{sec:conclusion}

We presented ReLoco, a framework that uses behavior-conditioned motion supervision to learn humanoid locomotion, fall recovery, and resumption of commanded motion within one proprioceptive policy. Robot posture conditions expert sampling and motion scoring across locomotion and recovery, with simulation-refined reference motions supporting policy learning. On the 64.5-kg Tiangong~3.0 humanoid, the deployed policy performs walking, running, autonomous get-up from arbitrary fallen postures, and resumption of commanded locomotion after falls, without on-robot policy fine-tuning.

\bibliographystyle{IEEEtran}
\bibliography{references}

@article{Schulman2017ppo,
  author  = {Schulman, John and Wolski, Filip and Dhariwal, Prafulla and Radford, Alec and Klimov, Oleg},
  title   = {Proximal Policy Optimization Algorithms},
  journal = {arXiv preprint arXiv:1707.06347},
  year    = {2017}
}

@inproceedings{Makoviychuk2021isaacgym,
  author = {Makoviychuk, Viktor and Wawrzyniak, Lukasz and Guo, Yunrong and Lu, Michelle and Storey, Kier and Macklin, Miles and Hoeller, David and Rudin, Nikita and Allshire, Arthur and Handa, Ankur and State, Gavriel},
  title = {{Isaac Gym}: High Performance {GPU} Based Physics Simulation for Robot Learning},
  booktitle = {Proc. Neural Inf. Process. Syst. Track Datasets Benchmarks},
  year = {2021},
  volume = {1},
  url = {https://datasets-benchmarks-proceedings.neurips.cc/paper_files/paper/2021/hash/28dd2c7955ce926456240b2ff0100bde-Abstract-round2.html}
}

@article{Peng2018deepmimic,
  author = {Peng, Xue Bin and Abbeel, Pieter and Levine, Sergey and van de Panne, Michiel},
  title = {{DeepMimic}: Example-Guided Deep Reinforcement Learning of Physics-Based Character Skills},
  journal = {ACM Transactions on Graphics},
  volume = {37},
  number = {4},
  pages = {143:1--143:14},
  year = {2018},
  doi = {10.1145/3197517.3201311},
  articleno = {143}
}

@article{Peng2021amp,
  author    = {Peng, Xue Bin and Ma, Ze and Abbeel, Pieter and Levine, Sergey and Kanazawa, Angjoo},
  title     = {{AMP}: Adversarial Motion Priors for Stylized Physics-Based Character Control},
  journal   = {ACM Transactions on Graphics},
  volume    = {40},
  number    = {4},
  pages     = {144:1--144:20},
  articleno = {144},
  numpages  = {20},
  year      = {2021},
  doi       = {10.1145/3450626.3459670}
}

@article{Peng2022ase,
  author    = {Peng, Xue Bin and Guo, Yunrong and Halper, Lina and Levine, Sergey and Fidler, Sanja},
  title     = {{ASE}: Large-Scale Reusable Adversarial Skill Embeddings for Physically Simulated Characters},
  journal   = {ACM Transactions on Graphics},
  volume    = {41},
  number    = {4},
  pages     = {94:1--94:17},
  articleno = {94},
  numpages  = {17},
  year      = {2022},
  doi       = {10.1145/3528223.3530110}
}

@inproceedings{Tessler2023calm,
  author    = {Tessler, Chen and Kasten, Yoni and Guo, Yunrong and Mannor, Shie and Chechik, Gal and Peng, Xue Bin},
  title     = {{CALM}: Conditional Adversarial Latent Models for Directable Virtual Characters},
  booktitle = {Proc. ACM SIGGRAPH},
  pages     = {1--9},
  year      = {2023},
  doi       = {10.1145/3588432.3591541}
}

@article{Won2022physicsvae,
  author    = {Won, Jungdam and Gopinath, Deepak and Hodgins, Jessica},
  title     = {Physics-Based Character Controllers Using Conditional {VAEs}},
  journal   = {ACM Transactions on Graphics},
  volume    = {41},
  number    = {4},
  pages     = {96:1--96:12},
  articleno = {96},
  numpages  = {12},
  year      = {2022},
  doi       = {10.1145/3528223.3530067}
}

@inproceedings{Luo2023phc,
  author    = {Luo, Zhengyi and Cao, Jinkun and Winkler, Alexander and Kitani, Kris and Xu, Weipeng},
  title     = {Perpetual Humanoid Control for Real-Time Simulated Avatars},
  booktitle = {Proc. IEEE/CVF Int. Conf. Comput. Vis. (ICCV)},
  pages     = {10895--10904},
  year      = {2023}
}

@article{Fussell2021supertrack,
  author  = {Fussell, Levi and Bergamin, Kevin and Holden, Daniel},
  title   = {{SuperTrack}: Motion Tracking for Physically Simulated Characters Using Supervised Learning},
  journal = {ACM Transactions on Graphics},
  volume  = {40},
  number  = {6},
  pages   = {1--13},
  year    = {2021},
  doi     = {10.1145/3478513.3480527}
}

@inproceedings{Rudin2022walk,
  author    = {Rudin, Nikita and Hoeller, David and Reist, Philipp and Hutter, Marco},
  title     = {Learning to Walk in Minutes Using Massively Parallel Deep Reinforcement Learning},
  booktitle = {Proc. 5th Conf. Robot Learn. (CoRL)},
  series    = {Proc. Mach. Learn. Res.},
  volume    = {164},
  pages     = {91--100},
  publisher = {PMLR},
  year      = {2022}
}

@inproceedings{Kumar2022adapting,
  author    = {Kumar, Ashish and Li, Zhongyu and Zeng, Jun and Pathak, Deepak and Sreenath, Koushil and Malik, Jitendra},
  title     = {Adapting Rapid Motor Adaptation for Bipedal Robots},
  booktitle = {Proc. IEEE/RSJ Int. Conf. Intell. Robots Syst. (IROS)},
  pages     = {1161--1168},
  year      = {2022},
  doi       = {10.1109/IROS47612.2022.9981091}
}

@article{Meduri2023biconmp,
  author = {Meduri, Avadesh and Shah, Paarth and Viereck, Julian and Khadiv, Majid and Havoutis, Ioannis and Righetti, Ludovic},
  title = {{BiConMP}: A Nonlinear Model Predictive Control Framework for Whole Body Motion Planning},
  journal = {IEEE Transactions on Robotics},
  volume = {39},
  number = {2},
  pages = {905--922},
  year = {2023},
  doi = {10.1109/TRO.2022.3228390}
}

@article{Jenelten2024dtc,
  author  = {Jenelten, Fabian and He, Junzhe and Farshidian, Farbod and Hutter, Marco},
  title   = {{DTC}: Deep Tracking Control},
  journal = {Science Robotics},
  volume  = {9},
  number  = {86},
  pages   = {eadh5401},
  year    = {2024},
  doi     = {10.1126/scirobotics.adh5401}
}

@article{Radosavovic2024humanoid,
  author  = {Radosavovic, Ilija and Xiao, Tete and Zhang, Bike and Darrell, Trevor and Malik, Jitendra and Sreenath, Koushil},
  title   = {Real-World Humanoid Locomotion with Reinforcement Learning},
  journal = {Science Robotics},
  volume  = {9},
  number  = {89},
  pages   = {eadi9579},
  year    = {2024},
  doi     = {10.1126/scirobotics.adi9579}
}

@article{Li2024cassie,
  author = {Li, Zhongyu and Peng, Xue Bin and Abbeel, Pieter and Levine, Sergey and Berseth, Glen and Sreenath, Koushil},
  title = {Reinforcement Learning for Versatile, Dynamic, and Robust Bipedal Locomotion Control},
  journal = {The International Journal of Robotics Research},
  volume = {44},
  number = {5},
  pages = {840--888},
  year = {2025},
  doi = {10.1177/02783649241285161}
}

@inproceedings{Masuda2024simtoreal,
  author = {Masuda, Shimpei and Takahashi, Kuniyuki},
  title = {Sim-to-Real Transfer of Compliant Bipedal Locomotion on Torque Sensor-Less Gear-Driven Humanoid},
  booktitle = {Proc. 22nd IEEE-RAS Int. Conf. Humanoid Robots (Humanoids)},
  pages = {1--8},
  year = {2023},
  doi = {10.1109/Humanoids57100.2023.10375181}
}

@inproceedings{Cheng2024exbody,
  author    = {Cheng, Xuxin and Ji, Yandong and Chen, Junming and Yang, Ruihan and Yang, Ge and Wang, Xiaolong},
  title     = {Expressive Whole-Body Control for Humanoid Robots},
  booktitle = {Proc. Robot. Sci. Syst. (RSS)},
  year      = {2024},
  doi       = {10.15607/RSS.2024.XX.107}
}

@inproceedings{Tang2024humanmimic,
  author    = {Tang, Annan and Hiraoka, Takuma and Hiraoka, Naoki and Shi, Fan and Kawaharazuka, Kento and Kojima, Kunio and Okada, Kei and Inaba, Masayuki},
  title     = {{HumanMimic}: Learning Natural Locomotion and Transitions for Humanoid Robot via Wasserstein Adversarial Imitation},
  booktitle = {Proc. IEEE Int. Conf. Robot. Autom. (ICRA)},
  pages     = {13107--13114},
  year      = {2024},
  doi       = {10.1109/ICRA57147.2024.10610449}
}

@inproceedings{Zhang2024whole,
  author    = {Zhang, Qiang and Cui, Peter and Yan, David and Sun, Jingkai and Duan, Yiqun and Han, Gang and Zhao, Wen and Zhang, Weining and Guo, Yijie and Zhang, Arthur and Xu, Renjing},
  title     = {Whole-Body Humanoid Robot Locomotion with Human Reference},
  booktitle = {Proc. IEEE/RSJ Int. Conf. Intell. Robots Syst. (IROS)},
  pages     = {11225--11231},
  year      = {2024},
  doi       = {10.1109/IROS58592.2024.10801451}
}

@inproceedings{He2024h2o,
  author    = {He, Tairan and Luo, Zhengyi and Xiao, Wenli and Zhang, Chong and Kitani, Kris and Liu, Changliu and Shi, Guanya},
  title     = {Learning Human-to-Humanoid Real-Time Whole-Body Teleoperation},
  booktitle = {Proc. IEEE/RSJ Int. Conf. Intell. Robots Syst. (IROS)},
  pages     = {8944--8951},
  year      = {2024},
  doi       = {10.1109/IROS58592.2024.10801984}
}

@inproceedings{Fu2025humanplus,
  author    = {Fu, Zipeng and Zhao, Qingqing and Wu, Qi and Wetzstein, Gordon and Finn, Chelsea},
  title     = {{HumanPlus}: Humanoid Shadowing and Imitation from Humans},
  booktitle = {Proc. 8th Conf. Robot Learn. (CoRL)},
  series    = {Proc. Mach. Learn. Res.},
  volume    = {270},
  pages     = {2828--2844},
  publisher = {PMLR},
  year      = {2025}
}

@inproceedings{He2025omnih2o,
  author    = {He, Tairan and Luo, Zhengyi and He, Xialin and Xiao, Wenli and Zhang, Chong and Zhang, Weinan and Kitani, Kris M. and Liu, Changliu and Shi, Guanya},
  title     = {{OmniH2O}: Universal and Dexterous Human-to-Humanoid Whole-Body Teleoperation and Learning},
  booktitle = {Proc. 8th Conf. Robot Learn. (CoRL)},
  series    = {Proc. Mach. Learn. Res.},
  volume    = {270},
  pages     = {1516--1540},
  publisher = {PMLR},
  year      = {2025}
}

@inproceedings{Cui2025terrain,
  author = {Cui, Wenhao and Li, Shengtao and Huang, Huaxing and Qin, Bangyu and Zhang, Tianchu and Han, Jinchao and Zheng, Liang and Tang, Ziyang and Hu, Chenxu and Yan, Ning and Chen, Jiahao and Jiang, Zheyuan},
  title = {Adapting Humanoid Locomotion over Challenging Terrain via Two-Phase Training},
  booktitle = {Proc. 8th Conf. Robot Learn. (CoRL)},
  series = {Proc. Mach. Learn. Res.},
  volume = {270},
  pages = {3439--3453},
  publisher = {PMLR},
  year = {2025}
}

@inproceedings{Chen2025smooth,
  author    = {Chen, Zixuan and He, Xialin and Wang, Yen-Jen and Liao, Qiayuan and Ze, Yanjie and Li, Zhongyu and Sastry, S. Shankar and Wu, Jiajun and Sreenath, Koushil and Gupta, Saurabh and Peng, Xue Bin},
  title     = {Learning Smooth Humanoid Locomotion through Lipschitz-Constrained Policies},
  booktitle = {Proc. IEEE/RSJ Int. Conf. Intell. Robots Syst. (IROS)},
  pages     = {4743--4750},
  year      = {2025},
  doi       = {10.1109/IROS60139.2025.11246506}
}

@inproceedings{Wang2025beamdojo,
  author    = {Wang, Huayi and Wang, Zirui and Ren, Junli and Ben, Qingwei and Huang, Tao and Zhang, Weinan and Pang, Jiangmiao},
  title     = {{BeamDojo}: Learning Agile Humanoid Locomotion on Sparse Footholds},
  booktitle = {Proc. Robot. Sci. Syst. (RSS)},
  year      = {2025},
  doi       = {10.15607/RSS.2025.XXI.068}
}

@inproceedings{Huang2025host,
  author    = {Huang, Tao and Ren, Junli and Wang, Huayi and Wang, Zirui and Ben, Qingwei and Wen, Muning and Chen, Xiao and Li, Jianan and Pang, Jiangmiao},
  title     = {Learning Humanoid Standing-Up Control across Diverse Postures},
  booktitle = {Proc. Robot. Sci. Syst. (RSS)},
  year      = {2025},
  doi       = {10.15607/RSS.2025.XXI.064}
}

@inproceedings{He2025humanup,
  author    = {He, Xialin and Dong, Runpei and Chen, Zixuan and Gupta, Saurabh},
  title     = {Learning Getting-Up Policies for Real-World Humanoid Robots},
  booktitle = {Proc. Robot. Sci. Syst. (RSS)},
  year      = {2025},
  doi       = {10.15607/RSS.2025.XXI.063}
}

@inproceedings{Chen2025hifar,
  author    = {Chen, Penghui and Wang, Yushi and Luo, Changsheng and Cai, Wenhan and Zhao, Mingguo},
  title     = {{HiFAR}: Multi-Stage Curriculum Learning for High-Dynamics Humanoid Fall Recovery},
  booktitle = {Proc. IEEE/RSJ Int. Conf. Intell. Robots Syst. (IROS)},
  pages     = {2908--2915},
  year      = {2025},
  doi       = {10.1109/IROS60139.2025.11245953}
}

@inproceedings{Gaspard2025frasa,
  author    = {Gaspard, Cl{\'e}ment and Duclusaud, Marc and Passault, Gr{\'e}goire and Daniel, M{\'e}lodie and Ly, Olivier},
  title     = {{FRASA}: An End-to-End Reinforcement Learning Agent for Fall Recovery and Stand Up of Humanoid Robots},
  booktitle = {Proc. IEEE Int. Conf. Robot. Autom. (ICRA)},
  pages     = {15994--16000},
  year      = {2025},
  doi       = {10.1109/ICRA55743.2025.11128664}
}

@inproceedings{Zhao2026ahc,
  author    = {Zhao, Yingnan and Wang, Xinmiao and Wang, Dewei and Liu, Xinzhe and Lu, Dan and Han, Qilong and Liu, Peng and Bai, Chenjia},
  title     = {Towards Adaptive Humanoid Control via Multi-Behavior Distillation and Reinforced Fine-Tuning},
  booktitle = {Proc. AAAI Conf. Artif. Intell.},
  volume    = {40},
  number    = {22},
  pages     = {18818--18826},
  year      = {2026},
  doi       = {10.1609/aaai.v40i22.38951}
}

@inproceedings{He2025asap,
  author    = {He, Tairan and Gao, Jiawei and Xiao, Wenli and Zhang, Yuanhang and Wang, Zi and Wang, Jiashun and Luo, Zhengyi and He, Guanqi and Sobanbabu, Nikhil and Pan, Chaoyi and Yi, Zeji and Qu, Guannan and Kitani, Kris and Hodgins, Jessica K. and Fan, Linxi and Zhu, Yuke and Liu, Changliu and Shi, Guanya},
  title     = {{ASAP}: Aligning Simulation and Real-World Physics for Learning Agile Humanoid Whole-Body Skills},
  booktitle = {Proc. Robot. Sci. Syst. (RSS)},
  year      = {2025},
  doi       = {10.15607/RSS.2025.XXI.066}
}

@inproceedings{He2025hover,
  author    = {He, Tairan and Xiao, Wenli and Lin, Toru and Luo, Zhengyi and Xu, Zhenjia and Jiang, Zhenyu and Kautz, Jan and Liu, Changliu and Shi, Guanya and Wang, Xiaolong and Fan, Linxi Jim and Zhu, Yuke},
  title     = {{HOVER}: Versatile Neural Whole-Body Controller for Humanoid Robots},
  booktitle = {Proc. IEEE Int. Conf. Robot. Autom. (ICRA)},
  pages     = {9989--9996},
  year      = {2025},
  doi       = {10.1109/ICRA55743.2025.11128549}
}

@inproceedings{Zhang2025gmp,
  author    = {Zhang, Haodong and Zhang, Liang and Chen, Zhenghan and Chen, Lu and Wang, Yue and Xiong, Rong},
  title     = {Natural Humanoid Robot Locomotion with Generative Motion Prior},
  booktitle = {Proc. IEEE/RSJ Int. Conf. Intell. Robots Syst. (IROS)},
  pages     = {6622--6629},
  year      = {2025},
  doi       = {10.1109/IROS60139.2025.11247682}
}

@inproceedings{Allshire2025videomimic,
  author    = {Allshire, Arthur and Choi, Hongsuk and Zhang, Junyi and McAllister, David and Zhang, Anthony and Kim, Chung Min and Darrell, Trevor and Abbeel, Pieter and Malik, Jitendra and Kanazawa, Angjoo},
  title     = {Visual Imitation Enables Contextual Humanoid Control},
  booktitle = {Proc. 9th Conf. Robot Learn. (CoRL)},
  series    = {Proc. Mach. Learn. Res.},
  volume    = {305},
  pages     = {794--815},
  publisher = {PMLR},
  year      = {2025}
}

@inproceedings{Li2025amo,
  author    = {Li, Jialong and Cheng, Xuxin and Huang, Tianshu and Yang, Shiqi and Qiu, Ri-Zhao and Wang, Xiaolong},
  title     = {{AMO}: Adaptive Motion Optimization for Hyper-Dexterous Humanoid Whole-Body Control},
  booktitle = {Proc. Robot. Sci. Syst. (RSS)},
  year      = {2025},
  doi       = {10.15607/RSS.2025.XXI.061}
}

@article{Colmenares2024retargeting,
  author  = {Baek, Sujin and Kim, Ahyeon and Choi, Jin-Young and Ha, Eunju and Kim, Jong-Wook},
  title   = {Human Motion Retargeting to a Full-Scale Humanoid Robot Using a Monocular Camera and Human Pose Estimation},
  journal = {International Journal of Control, Automation and Systems},
  volume  = {22},
  number  = {9},
  pages   = {2860--2870},
  year    = {2024},
  doi     = {10.1007/s12555-023-0686-y}
}

@inproceedings{Zhang2024smt,
  author    = {Zhang, Haodong and Chen, Zhike and Xu, Haocheng and Hao, Lei and Wu, Xiaofei and Xu, Songcen and Zhang, Zhensong and Wang, Yue and Xiong, Rong},
  title     = {Semantics-Aware Motion Retargeting with Vision-Language Models},
  booktitle = {Proc. IEEE/CVF Conf. Comput. Vis. Pattern Recognit. (CVPR)},
  pages     = {2155--2164},
  year      = {2024},
  doi       = {10.1109/CVPR52733.2024.00210}
}

@inproceedings{Araujo2026gmr,
  author = {Ara{\'u}jo, Jo{\~a}o Pedro and Ze, Yanjie and Xu, Pei and Wu, Jiajun and Liu, C. Karen},
  title = {Retargeting Matters: General Motion Retargeting for Humanoid Motion Tracking},
  booktitle = {Proc. IEEE Int. Conf. Robot. Autom. (ICRA)},
  year = {2026},
  pages = {6097--6104},
  doi = {10.1109/ICRA57385.2026.11697263}
}

@inproceedings{Chen2025motionimitation,
  author = {Chen, Zhenghan and Zhang, Haodong and Wang, Dongqi and Yu, Jiyu and Xu, Haocheng and Wang, Yue and Xiong, Rong},
  title = {A Whole-Body Motion Imitation Framework from Human Data for Full-Size Humanoid Robot},
  booktitle = {Proc. IEEE Int. Conf. Real-Time Comput. Robot. (RCAR)},
  year = {2025},
  doi = {10.1109/RCAR65431.2025.11139493},
  pages = {558--563}
}

@inproceedings{Wang2026xloco,
  author    = {Wang, Dewei and Wang, Xinmiao and Zhang, Chenyun and Shi, Jiyuan and Zhao, Yingnan and Bai, Chenjia and Li, Xuelong},
  title     = {{X-Loco}: Towards Generalist Humanoid Locomotion Control via Synergetic Policy Distillation},
  booktitle = {Proc. Robot. Sci. Syst. (RSS)},
  year      = {2026},
  address   = {Sydney, Australia},
  month     = {July},
  doi       = {10.15607/RSS.2026.XXII.022}
}

@article{Bjelonic2026pace,
  author    = {Bjelonic, Filip and Tischhauser, Fabian and Hutter, Marco},
  title     = {Towards Bridging the Gap: Systematic Sim-to-Real Transfer for Diverse Legged Robots},
  journal   = {The International Journal of Robotics Research},
  year      = {2026},
  doi       = {10.1177/02783649261459628},
  publisher = {SAGE Publications},
  note      = {Early access, doi: 10.1177/02783649261459628}
}

@article{liao2026beyondmimic,
  author = {Liao, Qiayuan and Truong, Takara E. and Huang, Xiaoyu and
             Gao, Yuman and Tevet, Guy and Sreenath, Koushil and
             Liu, C. Karen},
  title = {{BeyondMimic}: From Motion Tracking to Versatile Humanoid Control via Guided Diffusion},
  journal = {Science Robotics},
  volume = {11},
  number = {117},
  pages = {eadx8924},
  year = {2026},
  doi = {10.1126/scirobotics.adx8924}
}

@misc{BonesStudioSeed,
  author = {{Bones Studio}},
  title  = {{BONES-SEED}: Skeletal Everyday Embodiment Dataset},
  note   = {Dataset. Accessed: Nov. 10, 2026},
  url    = {https://huggingface.co/datasets/bones-studio/seed}
}

\end{document}